\documentclass[10pt,twocolumn,letterpaper]{article}

\usepackage{cvpr}
\usepackage{times}
\usepackage{epsfig}
\usepackage{graphicx}
\usepackage{amsmath}
\usepackage{amssymb}
\usepackage{mathtools}
\usepackage{dsfont}
\usepackage{makecell}

\usepackage[pagebackref=true,breaklinks=true,letterpaper=true,colorlinks,bookmarks=false]{hyperref}

\cvprfinalcopy 

\def\cvprPaperID{****} 

\ifcvprfinal\fi
\begin{document}

\title{Hierarchical Label Inference for Video Classification}

\author{Nelson Nauata\\
Simon Fraser University\\
{\tt\small nnauata@sfu.ca}
\and
Jonathan Smith\\
Simon Fraser University\\
{\tt\small jws4@sfu.ca}
\and
Greg Mori\\
Simon Fraser University\\
{\tt\small mori@cs.sfu.ca}
}
\maketitle

\begin{abstract}
    Videos are a rich source of high-dimensional structured data, with a wide range
	of interacting components at varying levels of granularity. In order to
	improve understanding of unconstrained internet videos, it is important to consider the
	role of labels at separate levels of abstraction. In this paper, we consider the use of the Bidirectional Inference Neural Network (BINN) for performing graph-based inference in label space for the task of video classification.
	We take advantage of the inherent hierarchy between labels at increasing granularity.
	The BINN is evaluated on the first and second release of the YouTube-8M large scale multi-label video dataset.
	Our results demonstrate the effectiveness of BINN, achieving significant improvements against baseline models.
\end{abstract}

\section{Introduction}

The proliferation of large-scale video
datasets (\cite{hmdb}, \cite{ucf101}, \cite{youtube8m}, \cite{sports1m}),
coupled with increasingly powerful computational resources allow
for applications of learning on an unprecedented level. In particular, the task
of labelling videos is of relevance with the massive flow of
unlabelled user-uploaded video on social media. The complex, rich nature of video data strongly motivates the use of deep learning, and
has seen measurable success in recent applications of video classification,
captioning, and question answering (\cite{lrcn}, \cite{videoCap}, \cite{movieQA}).

\begin{figure}
\centering
\includegraphics[width=0.5\textwidth]{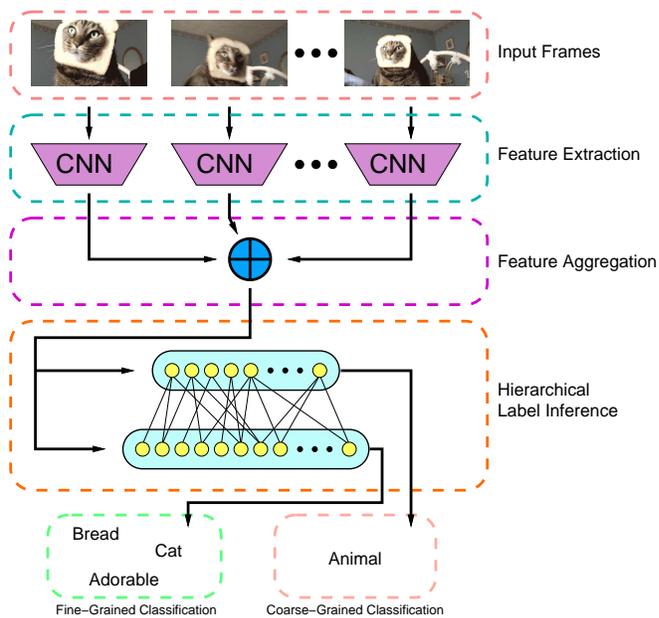}
\caption{Diagram of proposed model for performing video label inference. Each frame of a video is fed through a pre-trained CNN, followed by a mean pooling for temporal aggregation. Inference is performed in label space, and predictions are made at multiple levels of granularity.}
\label{fig1}
\end{figure}

Different labels such as \textit{outdoors} and \textit{mountain}, or 
\textit{beer} and \textit{Irish pub} have intrinsic
dependencies on each other that are difficult for standard deep learning methods
to model, as labels are generally assumed to be pairwise independent.
Graphical models have seen promising results on incorporating label-space
inference in image classification (\cite{ising}, \cite{inn}). In particular, the work in
\cite{inn} develops a Structured Inference Neural Network (SINN) and a 
Bidirectional Inference Neural Network (BINN) that performs
hierarchical inference for image labelling. However, an image is static stream
of data relative to video.

Models of temporal dependencies in sequential data have seen common use
in both computer vision and natural language processing.
In this paper, we investigate methods of incorporating recent methods
of label inference with convolutional neural networks, 
effectively combining spatial and hierarchichal information 
into a single end-to-end trainable network. Previous successful approaches to the
problem of video classification include Convolutional Neural Networks (CNNs) \cite{videoclassification},
Recurrent Neural Networks (RNNs) such as LSTMs (\cite{lrcn}, \cite{beyondshortsnippets}), 
Improved Dense Trajectories (IDT) \cite{idt}, and Histogram of Oriented Optical Flows (HOOF) \cite{hof}.
However, all of these previous models discount the hierarchical dependencies
between labels that could be leveraged to improve predictions -- this paper is an attempt at
resolving this disconnect. The explicit contribution of this paper is to extend the application of the BINN previously presented in \cite{inn} as a module for performing video classification equipped with structured prediction in label space. An overview of the model pipeline for a given video can be seen in Figure \ref{fig1}.


The evaluation of our model will follow in Section~\ref{sec:exp}. We calculate our results
on the two recent releases of the YouTube-8M large-scale video dataset \cite{youtube8m}. Our results will be compared against the current
state-of-the-art in \cite{youtube8m} and a baseline logistic regression classifier trained on the most recent release of YouTube-8M.

\section{Related Work}

\subsection{Label Relations}
By default, a standard video classification approach for a multi-label problem
will not model inherent dependencies between labels such as \textit{boat} and \textit{water}.
The incorporation of these label relations into classification problems with a large label
space has been shown to improve performance on multi-label classification tasks.
Graphical representations have been employed for modeling semantic relationships
within label space, such as the proposal of the HEX graph \cite{labelgraph} for modeling inter-label relations, or the Ising model \cite{ising}.

Applications of this concept to vision tasks has been seen in group activity recognition using
hierarchical relationships between actors and collective actions \cite{inferencemachine}. In particular, the SINN proposed in \cite{inn} leverages rich multi-label annotations in images for performing graph-based inference in label space.

\subsection{Video Understanding}
The use of machine learning for applications in video data has advanced in lockstep with the field's success with image data.
Recurrent neural networks (RNNs) have been recently utilized in a number of computer vision
tasks, notably for the purpose of video classification. Long Short Term Memory (LSTM) \cite{lstm}, an effective type
of RNN architecture, has seen frequent use in classification \cite{beyondshortsnippets}, 
video/image captioning \cite{lrcn}, and action recognition.
Beyond vision, the LSTM has seen success in Natural Language Processing (NLP) for machine anslation \cite{neuraltranslation}
and sentiment analysis \cite{sentiment}, and in speech recognition.

The use of RNNs to model relationships in object space has been explored, and focuses on
features from raw video input. In \cite{hierarchical}, dynamics between actors in a video scene
are jointly modeled by a set of LSTM units, and in \cite{structuralrnn}, spatio-temporal RNNs are
inferred from the relationships between objects in space.



\begin{figure*}
\centering
\includegraphics[width=\textwidth]{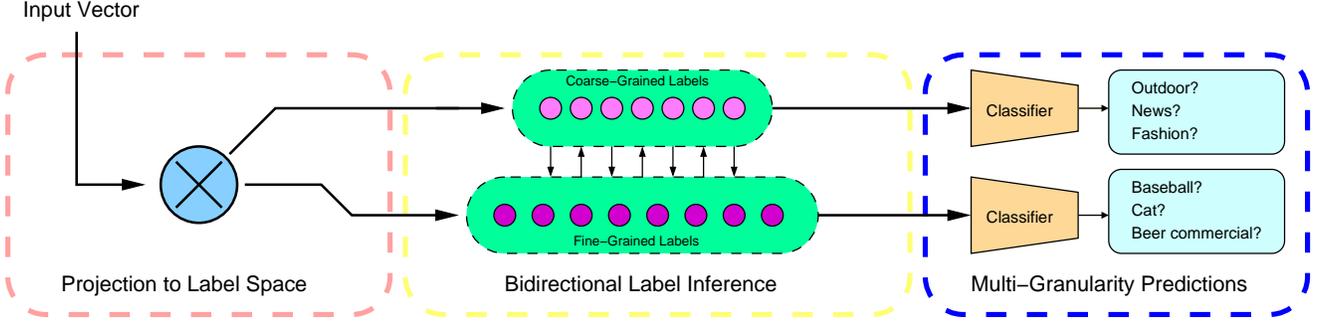}
\caption{Process of the BINN's prediction given an input activation vector. We begin with a fully connected layer to project the input vector to the dimensionality of the label space, as seen in eqn. 1. This is followed by the bidirectional inference performed on the fully connected label graph, where the vertices represent labels. Finally, the output activations at each concept layer are fed into a multi-label classifier.}
\label{fig2}
\end{figure*}

\section{Method}
\subsection{Bidirectional Inference Neural Network (BINN)}

The learning of structured label relations in \cite{inn} considers a hierarchical distribution of labels, where each level is defined as a concept layer and represents the degree of granularity for the label space in question. In the implementation for BINN, the labels are separated into a set of $m$ concept layers, with varying granularity. For example, a coarse-grained label for a scene could be \textit{outdoors} whereas a fine-grained label would be \textit{tree}. The intuition of the BINN is to treat concept layers
as separate timesteps of a bidirectional RNN, and propagate information and gradients between layers. In more explicit terms, a given input has labels $y_t$ at concept layer $t$.

Consider the input feature vector $x^i \in \mathbb{R}^D$. This vector will be treated as input separately for each concept layer.
The concept layers are represented as activations obtained by performing inference in the graph -- for each concept layer $t$, we have an activation vector $a_t \in \mathbb{R}^{k_t}$ associated with the labels at concept layer $t$, where $n_t$ is the number of labels at concept layer $t$. In order to perform inference, the dimension of the input $x^i$ must be regressed to the label space. Thus, the input $x^i_t$ for each concept layer t and each input feature vector $x^i$ is given by:
\begin{align}
	x^i_t = W_t * x^i + b_t
\end{align}
where $W_t \in \mathbb{R}^{n_t \times D}$ and $b_t \in \mathbb{R}^{n_t \times 1}$ are learnable parameters.

The bidirectional message passing consists of two steps: a top-down inference and a bottom-up inference. The former captures inter-layer and intra-layer label relations in the top-down direction computing intermediate activations represented by $\overrightarrow{a}_t$. The latter performs the same computation in the bottom-up direction and are represented by $\overleftarrow{a}_t$. 
Finally, aggregation parameters $\overrightarrow{U_t}$ and $\overleftarrow{U_t}$ are defined for combining both directions and obtaining $a_t$.
The entire formulation for a sample $x^i$ is the following:
\begin{align}
		\overrightarrow{a}_t &= \overrightarrow{V}_{t-1,t} \cdot \overrightarrow{a}_{t-1} + \overrightarrow{H}_t \cdot x^i_t + \overrightarrow{b}_t \\
		\overleftarrow{a}_t &= \overleftarrow{V}_{t+1,t} \cdot \overleftarrow{a}_{t+1} + \overleftarrow{H}_t \cdot x^i_t + \overleftarrow{b}_t \\
		a_t &= \overrightarrow{U}_t \cdot \overrightarrow{a}_t + \overleftarrow{U}_t \cdot \overleftarrow{a}_t + b_{a,t}
\end{align}
where $V_{i, j} \in \mathbb{R}^{n_j \times n_i}$, $H_i \in \mathbb{R}^{n_i \times n_i}$, $U_i \in \mathbb{R}^{n_i}$, and $b_{a, i}, b_{i} \in \mathbb{R}^{n_i}$ are all learnable parameters. The $V$ and $H$ weights capture the inter-layer and intra-layer dependencies, respectively. Since these parameters exhaust all pairwise relationships between labels, this step can be thought of as propagating activations across a fully-connected directed label graph.

In order to obtain class-specific probabilities for making label predictions, the activations are passed through a sigmoid function, yielding normalized activations denoted $\hat a_t$. To learn the layer parameters, the model will be trained end-to-end with backpropagation, minimizing the following logistic cross-entropy loss:
\begin{align}
		E(V) &= \sum\limits_{t=1}^{m} \sum\limits_{y=1}^{n_i} \bigg( \mathds{1} (y_t = y) \cdot \log \big( \sigma (\hat a_t) \big) \\
			 &+ \mathds{1} (y_t \neq y) \cdot \log \big( 1 - \sigma (\hat a_t) \big) \bigg)
\end{align}
where $E(V)$ denotes the error for a single video $V$. In training, we minimize the
sum $\sum\limits_i E(V_i)$ across each data batch $(V_1, V_2, \dots, V_B)$, where 
batch size $B$ is selected as a hyperparameter. A visual representation of the described model can be seen in Figure \ref{fig2}.

\subsection{Logistic Regression}
Our baseline model follows precisely the logistic regression model presented in \cite{youtube8m}. For a given video-level feature vector $x \in \mathbb{R}^D$, we define N entity specific parameters $w_\epsilon \in \mathbb{R}^{D+1}$, including the bias term. Thus, given an input sample $x \in \mathbb{R}^{D+1}$, we compute the probability of a certain entity $\epsilon$ as $p(\epsilon | x) = \sigma(w^T_\epsilon x)$. 
The optimization of $w_\epsilon$ parameters are performed by minimizing the loss function defined as:
\begin{align}
	\lambda||w_\epsilon||^2_2 + \sum^N_{i=1} \mathcal{L}(y_{i,e}, \sigma(w^T_\epsilon x))
\end{align}
where $\sigma(\cdot)$ is the sigmoid function.

\subsection{Video Representation}

The spatial feature representation provided by YT-8M and YT-8M V2 consists of pre-classification activations from the InceptionV3 network \cite{inceptionv3}, resulting in feature vectors of size 2048. According to \cite{youtube8m}, the spatial representations are decoded at 1 frame-per-second up to the first 360 seconds of video and are pre-processed using Principal Component Analysis (PCA), reducing the dimensionality to 1024 features per frame.

In order to summarize the information for an entire video sample, a simple aggregation approach was used. Since this paper's goal is to verify the validity of the BINN module, a simple averaging across all frame features is performed. The average summarizes a video's content into a single feature vector.

In addition to RGB frame features, the audio features are included in the YT-8M V2 dataset. The features, which are not included in YT-8M, are extracted using an acoustic CNN model as proposed in \cite{audio}. For the experimentation, the audio features will be averaged across a video and concatenated with the video-averaged RGB feature.
 
\subsubsection{Feature Normalization}
Before training, two steps of normalization are applied on the input samples. Two standard feature normalization techniques (Z-Normalization or PCA-Whitening) are applied to center the feature vectors and normalize to unit variance. Secondly, L2-normalization is performed on all features, which was seen to speed up model convergence. The effect of these preprocessing techniques on performance is explored in the results section.

\section{Experiments}
\label{sec:exp}

To prove its efficacy, the model will be empirically evaluated on YouTube-8M \cite{youtube8m}. The effect the exclusion/inclusion 
of portions of the proposed network and preprocessing steps will be investigated in order to justify each step of the method.

\subsection{YouTube-8M (YT-8M)}
The YouTube-8M dataset consists of approximately 8 million YouTube videos,
each annotated with 4800 Google Knowledge Graph \textit{entities}, functioning as
classes. With each entity label is associated up to 3 \textit{verticals} i.e. coarse-grained labels. The dataset is derived from roughly 500K hours of
untrimmed video, with an average of 1.8 labels per video. Each video is decoded
as a set of features extracted by passing the RGB frame through the InceptionV3 model from Szegedy \etal \cite{inceptionv3},
a deep CNN pre-trained on ImageNet \cite{imagenet}, and Principal Component Analysis (PCA) 
is applied to reduce feature dimension. The scale of this dataset in both label 
space and data space is unprecedented in the field of video datasets, surpassing 
previous benchmarks such as UCF-101 and Sports1M. Our results will be compared
against the published results on the validation set in \cite{youtube8m}.

\subsection{YouTube-8M V2 (YT-8M V2)}
The YouTube-8M V2 dataset represents the frame and audio features from approximately 7 million YouTube videos. The dataset is an updated version of YouTube-8M, with an increased number of labels per video and a smaller number of entities. On average, the videos in YT-8M V2 have 3.4 labels each, and there are only 4716 Google Knowledge Graph entities forming the label space. The preprocessing for this dataset is the same as YT-8M V1, but the audio features are also included, calculated using the CNN method in \cite{audio}. Our results on this dataset will be reported as well.

\subsection{Experiment Details}

The dataset labels were organized into a graph with two concept layers -- entities (i.e.\ fine-grained labels), and verticals (i.e.\ coarse-grained labels). We minimize the cross-entropy loss function using the Adam optimizer \cite{adam}. For all models, mini-batches of size 1024 were used, and a weight decay of $10^{-8}$ was applied. The logistic regression model was trained for 35k iterations with a learning rate of 0.01, and the BINN was trained for 90k iterations, starting with a learning rate of 0.001 with a decay factor of 0.1 at every 40k iterations. All models were implemented with the \textit{Caffe} deep learning framework \cite{caffe}, adapted from the implementation used in \cite{inn}.


\subsection{Results}

In Table \ref{table1}, results for YT-8M are presented on the validation set. The first section refers to the baseline models reported in \cite{youtube8m}. To ensure consistency with results published in \cite{youtube8m}, we trained their logistic regression baselines. From the results, the choice of feature normalization, Z-norm or PCA whitening, made little difference. On the other hand, the inclusion of L2-Normalization was crucial for duplicating the results. As can be verified from Table \ref{table1}, the best BINN model for YT-8M achieved state-of-art results with a marginally small improvement on the PERR and mAP metrics. However, using exclusively video-level features, an improvement of 2.19\% on mAP metric was seen against the leading baseline. The BINN also demonstrated measurable success on the PERR, Hit@1 and gAP metrics, with respective improvements of 4.51\%, 4.14\% and 6.72\%.

The results shown on Table \ref{table2} follows the same sequence as the previous results, save for for the inclusion of audio features available for YT-8M V2 and the lack of published results. The best results were obtained using RGB and audio features, with Z-norm and L2-Normalization. The most effective BINN model obtained 40.91\%, 72.27\%, 84.96\% and 79.29\% for mAP, PERR, Hit@1 and gAP respectively, which corresponds to improvements of 2.30\%, 2.99\%, 2.21\% and 3.46\%, against the leading baseline results.
\begin{table*}
\centering
\begin{tabular}{|l||c|c|c|c|}
    \hline Method & mAP & PERR & Hit@1 & gAP \\
    \Xhline{4\arrayrulewidth} LSTM \cite{youtube8m} & 26.60 & 57.30 & \textbf{64.50} & - \\
    \hline Mixture-of-Experts \cite{youtube8m} & 30.00 & 55.80 & 63.30 & -\\
    \hline Logistic Regression \cite{youtube8m} & 28.1 & 53.0 & 60.5 & -\\
    \Xhline {4\arrayrulewidth} Baseline + z-norm + l2-norm &  27.98 & 52.89& 60.34 & 49.04\\
    \hline Baseline + pca whitening + l2-norm & 27.74 & 52.82 & 60.31 & 48.78\\
    \hline BINN + z-norm + l2-norm  & \textbf{30.17} & \textbf{57.40} & 64.48 & \textbf{55.76}\\
    \hline BINN + pca whitening + l2-norm  & \textbf{30.17} & 57.39 & 64.49 & 55.73\\
    \hline
\end{tabular}
\caption{YouTube-8M (YT-8M) results for Mean Average Precision (mAP), Precision at Equal Recall Rate (PERR), Hit at 1 and gAP on the validation set.}
\label{table1}
\end{table*}

\begin{table*}
\centering
\begin{tabular}{|l||c|c|c|c|}
    \hline Method & mAP & PERR & Hit@1 & gAP \\
    \hline Baseline + rgb + z-norm + l2-norm & 36.84 & 64.38 & 78.62 & 70.31 \\
    \hline Baseline + rgb + pca whitening + l2-norm & 36.71 & 64.33 & 78.60 & 69.98\\
    \hline Baseline + rgb + audio + z-norm + l2-norm & 38.61 & 69.28 & 82.75 & 75.83\\
    \hline BINN + rgb + z-norm + l2-norm & 39.12 & 69.01 & 82.27 & 75.99\\
    \hline BINN + rgb + pca whitening + l2-norm& 38.42 & 69.03 & 82.31 & 75.98\\
    \hline BINN + rgb + audio + z-norm + l2-norm&  \textbf{40.91} & \textbf{72.27} & \textbf{84.96} & \textbf{79.29}\\
    \hline
\end{tabular}
\caption{YouTube-8M V2 (YT-8M V2) results for Mean Average Precision (mAP), Precision at Equal Recall Rate (PERR), Hit at 1 and gAP on the validation set.}
    \label{table2}
\end{table*}

\section{Conclusion}

A recently proposed method for graph-based inference was shown to achieve state-of-the-art results on a new large-scale video database. Via the modular design of the approach, it can be applied as an additional segment of model of larger scale. In the terms of future work, using video averaging as a temporal aggregation technique is naive -- a model that considers label  dependencies in time is a natural extension. The incorporation of a recurrent model like an LSTM or label inference in temporal space would achieve an end-to-end trainable example of such an architecture. Additionally, the model could be applied to any other dataset with labels that can be organized into a hierarchy, and is not restricted to YouTube-8M.

{\small
\bibliographystyle{ieee}
\bibliography{egbib}

\begin{thebibliography}{10}\itemsep=-1pt

\bibitem{youtube8m}
S.~Abu{-}El{-}Haija, N.~Kothari, J.~Lee, P.~Natsev, G.~Toderici,
  B.~Varadarajan, and S.~Vijayanarasimhan.
\newblock Youtube-8m: {A} large-scale video classification benchmark.
\newblock {\em CoRR}, abs/1609.08675, 2016.

\bibitem{hof}
R.~Chaudhry, A.~Ravichandran, G.~Hager, and R.~Vidal.
\newblock Histograms of oriented optical flow and binet-cauchy kernels on
  nonlinear dynamical systems for the recognition of human actions.
\newblock In {\em 2009 IEEE Conference on Computer Vision and Pattern
  Recognition}, pages 1932--1939, June 2009.

\bibitem{labelgraph}
J.~Deng, N.~Ding, Y.~Jia, A.~Frome, K.~Murphy, S.~Bengio, Y.~Li, H.~Neven, and
  H.~Adam.
\newblock {\em Large-Scale Object Classification Using Label Relation Graphs},
  pages 48--64.
\newblock Springer International Publishing, Cham, 2014.

\bibitem{inferencemachine}
Z.~Deng, A.~Vahdat, H.~Hu, and G.~Mori.
\newblock Structure inference machines: Recurrent neural networks for analyzing
  relations in group activity recognition.
\newblock {\em CoRR}, abs/1511.04196, 2015.

\bibitem{ising}
N.~Ding, J.~Deng, K.~Murphy, and H.~Neven.
\newblock Probabilistic label relation graphs with ising models.
\newblock {\em CoRR}, abs/1503.01428, 2015.

\bibitem{lrcn}
J.~Donahue, L.~A. Hendricks, S.~Guadarrama, M.~Rohrbach, S.~Venugopalan,
  K.~Saenko, and T.~Darrell.
\newblock Long-term recurrent convolutional networks for visual recognition and
  description.
\newblock {\em CoRR}, abs/1411.4389, 2014.

\bibitem{audio}
S.~Hershey, S.~Chaudhuri, D.~P.~W. Ellis, J.~F. Gemmeke, A.~Jansen, C.~Moore,
  M.~Plakal, D.~Platt, R.~A. Saurous, B.~Seybold, M.~Slaney, R.~Weiss, and
  K.~Wilson.
\newblock Cnn architectures for large-scale audio classification.
\newblock In {\em International Conference on Acoustics, Speech and Signal
  Processing (ICASSP)}. 2017.

\bibitem{lstm}
S.~Hochreiter and J.~Schmidhuber.
\newblock Long short-term memory.
\newblock {\em Neural Comput.}, 9(8):1735--1780, Nov. 1997.

\bibitem{inn}
H.~Hu, G.~Zhou, Z.~Deng, Z.~Liao, and G.~Mori.
\newblock Learning structured inference neural networks with label relations.
\newblock {\em CoRR}, abs/1511.05616, 2015.

\bibitem{hierarchical}
M.~S. Ibrahim, S.~Muralidharan, Z.~Deng, A.~Vahdat, and G.~Mori.
\newblock Hierarchical deep temporal models for group activity recognition.
\newblock {\em CoRR}, abs/1607.02643, 2016.

\bibitem{structuralrnn}
A.~Jain, A.~R. Zamir, S.~Savarese, and A.~Saxena.
\newblock Structural-rnn: Deep learning on spatio-temporal graphs.
\newblock {\em CoRR}, abs/1511.05298, 2015.

\bibitem{caffe}
Y.~Jia, E.~Shelhamer, J.~Donahue, S.~Karayev, J.~Long, R.~Girshick,
  S.~Guadarrama, and T.~Darrell.
\newblock Caffe: Convolutional architecture for fast feature embedding.
\newblock {\em arXiv preprint arXiv:1408.5093}, 2014.

\bibitem{sports1m}
A.~Karpathy, G.~Toderici, S.~Shetty, T.~Leung, R.~Sukthankar, and L.~Fei-Fei.
\newblock Large-scale video classification with convolutional neural networks.
\newblock In {\em CVPR}, 2014.

\bibitem{videoclassification}
A.~Karpathy, G.~Toderici, S.~Shetty, T.~Leung, R.~Sukthankar, and L.~Fei-Fei.
\newblock Large-scale video classification with convolutional neural networks.
\newblock In {\em CVPR}, 2014.

\bibitem{adam}
D.~P. Kingma and J.~Ba.
\newblock Adam: {A} method for stochastic optimization.
\newblock {\em CoRR}, abs/1412.6980, 2014.

\bibitem{hmdb}
H.~Kuehne, H.~Jhuang, E.~Garrote, T.~Poggio, and T.~Serre.
\newblock {HMDB}: a large video database for human motion recognition.
\newblock In {\em Proceedings of the International Conference on Computer
  Vision (ICCV)}, 2011.

\bibitem{beyondshortsnippets}
J.~Y. Ng, M.~J. Hausknecht, S.~Vijayanarasimhan, O.~Vinyals, R.~Monga, and
  G.~Toderici.
\newblock Beyond short snippets: Deep networks for video classification.
\newblock {\em CoRR}, abs/1503.08909, 2015.

\bibitem{imagenet}
O.~Russakovsky, J.~Deng, H.~Su, J.~Krause, S.~Satheesh, S.~Ma, Z.~Huang,
  A.~Karpathy, A.~Khosla, M.~S. Bernstein, A.~C. Berg, and F.~Li.
\newblock Imagenet large scale visual recognition challenge.
\newblock {\em CoRR}, abs/1409.0575, 2014.

\bibitem{ucf101}
K.~Soomro, A.~R. Zamir, and M.~Shah.
\newblock {UCF101:} {A} dataset of 101 human actions classes from videos in the
  wild.
\newblock {\em CoRR}, abs/1212.0402, 2012.

\bibitem{neuraltranslation}
I.~Sutskever, O.~Vinyals, and Q.~V. Le.
\newblock Sequence to sequence learning with neural networks.
\newblock {\em CoRR}, abs/1409.3215, 2014.

\bibitem{inceptionv3}
C.~Szegedy, V.~Vanhoucke, S.~Ioffe, J.~Shlens, and Z.~Wojna.
\newblock Rethinking the inception architecture for computer vision.
\newblock {\em CoRR}, abs/1512.00567, 2015.

\bibitem{movieQA}
M.~Tapaswi, Y.~Zhu, R.~Stiefelhagen, A.~Torralba, R.~Urtasun, and S.~Fidler.
\newblock Movieqa: Understanding stories in movies through question-answering.
\newblock {\em CoRR}, abs/1512.02902, 2015.

\bibitem{videoCap}
S.~Venugopalan, H.~Xu, J.~Donahue, M.~Rohrbach, R.~J. Mooney, and K.~Saenko.
\newblock Translating videos to natural language using deep recurrent neural
  networks.
\newblock {\em CoRR}, abs/1412.4729, 2014.

\bibitem{idt}
H.~Wang and C.~Schmid.
\newblock Action recognition with improved trajectories.
\newblock In {\em 2013 IEEE International Conference on Computer Vision}, pages
  3551--3558, Dec 2013.

\bibitem{sentiment}
J.~Xu, D.~Chen, X.~Qiu, and X.~Huang.
\newblock Cached long short-term memory neural networks for document-level
  sentiment classification.
\newblock {\em CoRR}, abs/1610.04989, 2016.

\end{thebibliography}
}

\end{document}